# Weak Form Theory-guided Neural Network (TgNN-wf) for Deep Learning of Subsurface Single and Two-phase Flow

Rui Xu[1], Dongxiao Zhang[2,*], Miao Rong[1], Nanzhe Wang[3]

[1] Intelligent Energy Lab, Peng Cheng Laboratory, Guangdong, China.

[2] School of Environmental Science and Engineering, Southern University of Science and Technology, Guangdong, China.

[3] College of Engineering, Peking University, Beijing, China.

[*] Corresponding author: Dongxiao Zhang (zhangdx@sustech.edu.cn)

**Key Points:**

- Weak form constraints are incorporated into a fully connected neural network to predict future responses.

- Domain decomposition is used with locally defined test functions to reduce computational cost and to capture local discontinuity.

- Our model shows improved accuracy and robustness to noises compared to strong form theory-guided neural networks.

**Abstract**

Deep neural networks (DNNs) are widely used as surrogate models in geophysical applications; incorporating theoretical guidance into DNNs has improved the generalizability. However, most of such approaches define the loss function based on the strong form of conservation laws (via partial differential equations, PDEs), which is subject to deteriorated accuracy when the PDE has high order derivatives or the solution has strong discontinuities. Herein, we propose a weak form theory-guided neural network (TgNN-wf), which incorporates the weak form formulation of the PDE into the loss function combined with data constraint and initial and boundary conditions regularizations to tackle the aforementioned difficulties. In the weak form, high order derivatives in the PDE can be transferred to the test functions by performing integration-by-parts, which reduces computational error. We use domain decomposition with locally defined test functions, which captures local discontinuity effectively. Two numerical cases demonstrate the superiority of the proposed TgNN-wf over the strong form TgNN, including the hydraulic head prediction for unsteady-state 2D single-phase flow problems and the saturation profile prediction for 1D two-phase flow problems. Results show that TgNN-wf consistently has higher accuracy than TgNN, especially when strong discontinuity in the solution is present. TgNN-wf also trains faster than TgNN when the number of integration subdomains is not too large (<10,000). Moreover, TgNN-wf is more robust to noises. Thus, the proposed TgNN-wf paves the way for which a variety of deep learning problems in the small data regime can be solved more accurately and efficiently.

**1 Introduction**

The study of single and two-phase flow problems has important applications in subsurface geophysical problems, such as hydrocarbon resources recovery, contaminant transport, and groundwater flow and transport (Almasri & Kaluarachchi, 2005; Aziz & Settari, 1979; Blunt, 2001). These problems can generally be described by partial differential equations (PDEs) constructed from the conservation laws. Traditionally, PDEs can be solved numerically by finite difference, finite element, or finite volume methods (Aziz & Settari, 1979). These methods have been well-established and are known for their robustness and flexibility. However, in regards to long-time prediction, uncertainty quantification, or inverse modeling, in which hundreds of thousands of evaluation of these numerical models are required, computational efficiency becomes a major bottleneck (Mo et al., 2019; Zhu et al., 2019). Moreover, expensive computational mesh generation and refinement is usually required for these methods (Sirignano & Spiliopoulos, 2018). If the problem setup is changed (e.g., geometry of the simulation domain, boundary conditions, or medium properties), the meshes need to be regenerated to accommodate such a change.

To compensate for the deficiencies of numerical models, surrogate models have been used to facilitate the evaluation process, for example, wavelet analysis (Daubechies, 1992), Gaussian kernels (Hangelbroek & Ron, 2010), spline interpolation (Scarpiniti et al., 2013), and neural network approximation (Jo et al., 2019; LeCun et al., 2015). Deep neural networks (DNNs), in particular, have dramatically evolved over the past decades due to advances in theories and computational power. They serve as powerful nonlinear function approximators and have shown unprecedented performance in the fields of image recognition (He et al., 2016), natural language processing (Collobert & Weston, 2008), and autonomous driving (Sallab et al., 2017), where a large amount of training data is readily available. DNNs operate well in the big data regime, however, in many real-world geophysical applications where the acquisition of data is usually expensive or time-consuming, DNNs trained in the small data regime usually suffer from a lack of robustness and accuracy (Karpatne et al., 2017). In recent years, theoretical guidance and physical laws (in the form of PDEs) have been integrated into DNNs to further constrain the solution space. Instead of solving PDEs on a computational mesh, the residual of PDEs is constrained on chosen collocation points. Therefore, such approaches are completely mesh-free. Scientific problems in the areas of turbulence modeling, quantum mechanics, biology, climate science, and hydrology have been addressed with success using this approach. Remarkable improvement has been observed compared to DNNs trained solely on scarce data. Karpatne et al. (2017) formally conceptualized the framework of theory-guided data science (TGDS) and described several approaches for integrating domain knowledge into the machine learning framework. Raissi and coworkers proposed a physics-informed neural network (PINN) in which PDE constraints were incorporated into the loss functions of DNNs (Raissi et al., 2019; Raissi & Karniadakis, 2018). They studied both forward and inverse problems, demonstrating the robustness of PINN when noises were introduced into the training data. Later, variations of PINN have been used for a variety of forward and inverse problems, which require the solution of PDEs (Jagtap et al., 2020; Pang et al., 2018). Recently, Wang et al. (2020) proposed a theory-guided neural network (TgNN) which incorporates physical laws, expert knowledge, and engineering control into the loss functions, demonstrating its robustness in solving 2D single phase flow problems. Furthermore, some works showed the capability of theory-informed neural networks to solve PDEs directly by minimizing the loss function constructed only from PDEs without the need of labeled data (e.g. the so-called 'simulator-free' approach (Karumuri et al., 2020; Wang et al., 2020)). Most of the work defines the loss function based on the strong form of the conservation laws (PDE), which is subject to deteriorated accuracy, efficiency, or robustness when the PDE has high order derivatives or the solution has strong discontinuities/singularities, or when the available training data is noisy. To compensate for that, weak form formulation of the mathematical models has been adopted to construct the loss functions.

Weak form is formulated by multiplying the PDE with certain test functions and integrating over the simulation domain. It has several advantages compared to the strong form:

1) The order of derivatives in the PDE can be effectively reduced by performing integration-by-parts, avoiding the repeated evaluation of high order derivatives, whichcan be time-consuming and error-prone.

2) Rather than evaluating the PDE at individual collocation points, weak form evaluates the integral over either the whole domain or a certain number of subdomains (we call them collocation regions). Therefore, integrating and averaging over the subdomain can reduce the effect of noises in the training data.

3) The size of the collocation region can be tailored to capture local discontinuity or fluctuation with the flexibility of performing local domain refinement.

Kharazmi et al. (2019) formulated a variational PINN (VPINN) as an improvement of the previously developed PINN by incorporating the weak form loss function into a fully connected neural network (FCNN). A Petrov-Galerkin method was used to evaluate the integrals over the entire computational domain, taking the FCNN as the trial function and polynomials or trigonometric functions as test functions. Improved accuracy was observed for VPINN when solving problems with high-order derivatives. Later, they extended their work using domain decomposition and evaluated the integrals on subdomains (Kharazmi et al., 2020). The efficiency and accuracy of their improved VPINN was shown on several numerical examples with non-smooth solutions. Bao et al. (2020) used weak adversarial networks (WANs) to solve PDEs by taking the trial and test functions as separate neural networks, which were trained collectively in the process. The accuracy and efficiency of their proposed approach was demonstrated on a variety of inverse problems. E & Yu (2017) proposed a Deep Ritz method to solve variational problems by taking the test functions to be the strong form residuals, and illustrated their method on solving high dimensional PDEs and eigenvalue problems.

Commonly, the loss function of the neural network is the summation of several parts, including contributions from either strong or weak form residual of the conservation laws, the data mismatch, and initial (IC) and boundary conditions (BC), etc. All work mentioned above tuned the weight of each term in the loss function either by experience or by trial and error, which might be time-consuming and be subject to human experience. The weights are also not necessarily the optimum choice. Furthermore, most of the studies focus on low-dimension steady-state problems. The introduction of time evolution increases the input dimension and requires a more delicate test function choice and domain decomposition. Herein, we formulate a weak form TgNN (TgNN-wf) framework by taking a FCNN as the base and incorporating weak form constraint, data mismatch, and IC/BC regularizations into the loss function. The weights of the terms in the loss function are optimized by reformulating the original loss minimization problem into a Lagrangian duality problem (Fioretto et al., 2020; Rong et al., 2020). We construct test functions tailored to the problems, which effectively reduces the computational cost. Domain decomposition is used to reduce the number of Gaussian quadrature points needed for integral evaluation when high dimensional input space is considered and to better capture the discontinuity in the solution space. The robustness of our proposed approach is demonstrated through two numerical examples, which includes the hydraulic head prediction for 2D unsteady-state single phase flow problem and the saturation profile prediction for 1D Buckley-Leverett two-phase flow problem. We perform detailed comparison between TgNN-wf and TgNN in terms of accuracy, computational efficiency, and robustness to noises in the training data.

The remainder of the paper is organized as follows. In Section 2, we introduce the structure of the basic DNN and the formulation of our proposed TgNN-wf. In Section 3, we show the two numerical cases based on which detailed comparison between TgNN and TgNN-wf are made. Conclusions follow in Section 4.

## 2 Methodology

### 2.1 Deep Neural Network (DNN)

DNN is a powerful nonlinear function approximator that has evolved dramatically in the past decades. It can learn the complex projection mapping between the inputs and outputs using a relatively simple network structure. Usually, a DNN is composed of an input layer, several hidden layers, and an output layer. For a DNN with $L$ hidden layers, the feedforward formulation can be written as:

$$\mathbf{T}^{(i)} = \sigma_i\big(\mathbf{W}^{(i)}\mathbf{X}^{(i-1)} + \mathbf{b}^{(i)}\big), \qquad i = 1, 2, \cdots, L+1 \tag{1}$$

where $\mathbf{T}^{(i)}$ is the output of the $i$th layer, or the input of the $(i+1)$th layer, and $\mathbf{T}^{(L+1)}$ corresponds to the output vector. $\mathbf{W}^{(i)}$ and $\mathbf{b}^{(i)}$ are the weights and biases matrices of the $i$th layer, $\mathbf{X}^{(i-1)}$ is the neuron values of the $(i-1)$th layer, and $\mathbf{X}^{(0)}$ corresponds to the input vector. $\sigma_i$ is the activation function of the $i$th layer. Common choices of activation function include hyperbolic tangent (Tanh), Sigmoid, and Rectified Linear Unit (ReLU) and its variations (Goodfellow et al., 2016). During the training process, the following mean squared error (MSE) loss function regarding the data mismatch is minimized:

$$L(\mathbf{W}, \mathbf{b}) = \frac{1}{N_d} \sum_{i=1}^{N_d} (u_i^{NN}(\mathbf{X}_i; \mathbf{W}, \mathbf{b}) - u_i)^2 \tag{2}$$

where $u$ is the true solution, and $u^{NN}$ is the neural network approximation of the solution $u$; $N_d$ is the total number of training data points. Eq. 2 can be minimized to obtain the network parameters $(\mathbf{W}, \mathbf{b})$ using widely adopted algorithms, such as stochastic gradient decent or Adam (Goodfellow et al., 2016).

### 2.2 Theory-guided neural network (TgNN)

We first briefly introduce the structure of TgNN (Wang et al., 2020) based on which TgNN-wf is developed. Here we consider initial-boundary value problems for which the governing equations can be written as:

$$L_p u(\mathbf{x}, t) = f(\mathbf{x}, t), \qquad \mathbf{x} \in \Omega, t \in (0, T] \tag{3.1}$$

$$I u(\mathbf{x}, 0) = g(\mathbf{x}), \qquad \mathbf{x} \in \Omega \tag{3.2}$$

$$Bu(\mathbf{x}, t) = h(\mathbf{x}), \qquad \mathbf{x} \in \partial\mathbf{\Omega}, t \in (0, T] \tag{3.3}$$

Eq. 3.1 is defined over the spatial and temporal domain $\mathbf{\Omega} \subset \mathbb{R}^d$, $t \in (0, T)$ with spatial boundaries $\partial\mathbf{\Omega}$. $L_p$ is a differential operator that contains the partial derivatives of $u$. Operators $I$ and $B$ define the initial and boundary conditions, respectively. In TgNN, the loss function is composed of four general parts, including the data mismatch ($R_{data}$), the strong form (PDE) residual ($R_{sf}$), and IC/BC regularization ($R_{IC}$ and $R_{BC}$) (Wang et al., 2020):

$$R_{data} = \frac{1}{N_d} \sum_{i=1}^{N_d} [u_i^{NN}(\mathbf{x}_i, t_i; \mathbf{W}, \mathbf{b}) - u_i(\mathbf{x}_i, t_i)]^2 \tag{4.1}$$

$$R_{sf} = \frac{1}{N_c} \sum_{i=1}^{N_c} [L_p u_i^{NN}(\mathbf{x}_i, t_i; \mathbf{W}, \mathbf{b}) - f(\mathbf{x}_i, t_i; \mathbf{W}, \mathbf{b})]^2 \tag{4.2}$$

$$R_{IC} = \frac{1}{N_{IC}} \sum_{i=1}^{N_{IC}} [u_i^{NN}(\mathbf{x}_i, 0; \mathbf{W}, \mathbf{b}) - u_i(\mathbf{x}_i, 0)]^2 \tag{4.3}$$

$$R_{BC} = \frac{1}{N_{BC}} \sum_{i=1}^{N_{BC}} [u_i^{NN}(\partial\mathbf{\Omega}_i, t_i; \mathbf{W}, \mathbf{b}) - u_{BC}(\partial\mathbf{\Omega}_i, t_i)]^2 \tag{4.4}$$

where $N_d$ is the total number of training data points, $N_c$ is the number of collocation points for PDE residual evaluation, and $N_{IC}$ and $N_{BC}$ are the numbers of collocation points for the evaluation of initial and boundary conditions, respectively. Note that here we assume the boundary condition is of Dirichlet type, while for Neuman boundaries, $R_{BC}$ can also be easily derived. The loss function for TgNN is then defined as the weighted summation of the four residual components:

$$L_{TgNN}(\mathbf{W}, \mathbf{b}) = \lambda_d R_{data} + \lambda_f R_{sf} + \lambda_{IC} R_{IC} + \lambda_{BC} R_{BC}, \tag{5}$$

where $\lambda_d$, $\lambda_f$, $\lambda_{IC}$, and $\lambda_{BC}$ are the weights of each residual term. These weights are commonly tuned by hand, based on either experience or trial and error and remain constant during the training process. There is no systematic analysis of weight determination or optimization in the literature. Different weight choices have been reported in previous works (Kharazmi et al., 2019; Raissi et al., 2019; Wang et al., 2020), which is a function of the problem to be solved, the network structure, and the magnitude of each residual term, etc. Given Eq. 5, the model training process can be formulated as the following minimization problem: $\min_{\mathbf{W},\mathbf{b}} L_{TgNN}(\mathbf{W},\mathbf{b})$.

2.3 Weak form theory-guided neural network (TgNN-wf)

To reformulate the PDE into the corresponding weak form, we first decompose the numerical domain $\mathbf{\Omega} \times (0,T]$ into $N_w$ space-time collocation regions $\Delta\mathbf{\Omega}_i \times \Delta t_i$. On each collocation region, we multiply both sides of Eq. 3.1 by a locally defined test function $\omega$ and integrate over the subdomain:

$$\int_{\Delta\mathbf{\Omega}_i \times \Delta t_i} L_p u(\mathbf{x},t) \omega_i(\mathbf{x},t) d\mathbf{\Omega} dt = \int_{\Delta\mathbf{\Omega}_i \times \Delta t_i} f(\mathbf{x},t) \omega_i(\mathbf{x},t) d\mathbf{\Omega} dt. \tag{6}$$

The residual of the neural network approximation of Eq. 6 can be written as:

$$r_{wf_i} = \int_{\Delta\mathbf{\Omega}_i \times \Delta t_i} [L_p u^{NN}(\mathbf{x},t) \omega_i(\mathbf{x},t) - f(\mathbf{x},t) \omega_i(\mathbf{x},t)] d\mathbf{\Omega} dt. \tag{7}$$

The trial function in our weak form formulation is the global nonlinear function approximated by the neural network. Different choices of test functions have been reported in the literature. They can either be a single specifically-designed function that is defined locally on each subdomain (E & Yu, 2017; Reinbold et al., 2020), or a set of orthogonal functions (e.g. polynomial or trigonometric functions (Kharazmi et al., 2019; Kharazmi et al., 2020)) used to approximate the real solution, or they can be approximated by another neural network that is trained collectively in the process (Bao et al., 2020). Here we define test functions the first way, which will be shown later in Section 3.

The partial derivative evaluation in the neural network can be achieved by automatic differentiation (Paszke et al., 2017). However, when high order derivatives are present in $L_p$, automatic differentiation might be time-consuming and error-prone. The biggest advantage of the weak form formulation is that by performing integration-by-parts for Eq. 7, the high order differential operators in $L_p$ can be transferred from $u$ to $\omega$, therefore reducing computational error and cost since the derivatives of $\omega$ can be calculated analytically. Moreover, by performing domain decomposition, we could change the size of the collocation region or perform local domain refinement to better capture discontinuity in the solution.

The loss function of TgNN-wf is obtained by replacing the strong form residual $R_{sf}$ in Eq. 5 with its weak form counterpart,

$$R_{wf} = \frac{1}{N_c} \sum_{i=1}^{N_w} r_{wf_i}^2, \tag{8}$$

where $N_c$ is the total number of collocation regions. We get

$$L_{TgNN-wf}(\mathbf{W}, \mathbf{b}) = \lambda_d R_{data} + \lambda_f R_{wf} + \lambda_{IC} R_{IC} + \lambda_{BC} R_{BC}, \tag{9}$$

where $R_{data}$, $R_{IC}$, and $R_{BC}$ take the same form with TgNN. Similarly, the solution of the PDE is transformed into the minimization problem: $\min_{\mathbf{W},\mathbf{b}} L_{TgNN-wf}(\mathbf{W}, \mathbf{b})$.

2.4 Lagrangian duality

Instead of using constant weights $\lambda$ in Eq. 5 and 9 in the training process, here we treat the weights as hyperparameters and optimize them during the training process by reformulating the original minimization problem into a Lagrangian duality problem (Rong et al., 2020). Here we keep $\lambda_d = \lambda_{IC} = \lambda_{BC} = 1$ and only optimize the value of $\lambda_f$. This is because $R_{data}$, $R_{IC}$, and $R_{BC}$ are all evaluated on training data or on collocation points and have consistent physical unit and dimension, therefore, their contribution can be treated equally. On the other hand, $R_{wf}$ is evaluated in the functional form and not necessarily in the same physical unit and dimension with the former three. $\lambda_f$ can therefore be treated as a conversion factor to scale $R_{wf}$ to comparable magnitude and dimension with the former three. As such, it is noted that the weights may not be dimensionless. To illustrate the Lagrangian duality formulation, we take TgNN-wf as an example, wherein similar workflow can be applied to TgNN. For TgNN-wf, the original minimization problem can be written as:

$$\min_{\mathbf{W},\mathbf{b}} R_{data} + R_{IC} + R_{BC}, \quad \text{s.t. } R_{wf} = 0 \tag{10}$$

We then define the generalized Lagrangian equation in the unconstrained form:

$$L_d(\lambda_f, \mathbf{W}, \mathbf{b}) = R_{data} + R_{IC} + R_{BC} + \lambda_f R_{wf}. \tag{11}$$

The Lagrangian duality problem can then be defined as (Rong et al., 2020):

$$\max_{\lambda_f} \min_{\mathbf{W},\mathbf{b}} L_d(\lambda_f, \mathbf{W}, \mathbf{b}), \quad \text{s.t. } \lambda_f > 0. \tag{12}$$

In the training process, $\lambda_f$ is updated in each iteration after the optimization of $\mathbf{W}, \mathbf{b}$. Different optimization algorithms and learning rates can be applied to the two processes based on the problem to be solved. **Fig. 1** illustrates the overall structure of the TgNN-wf framework.

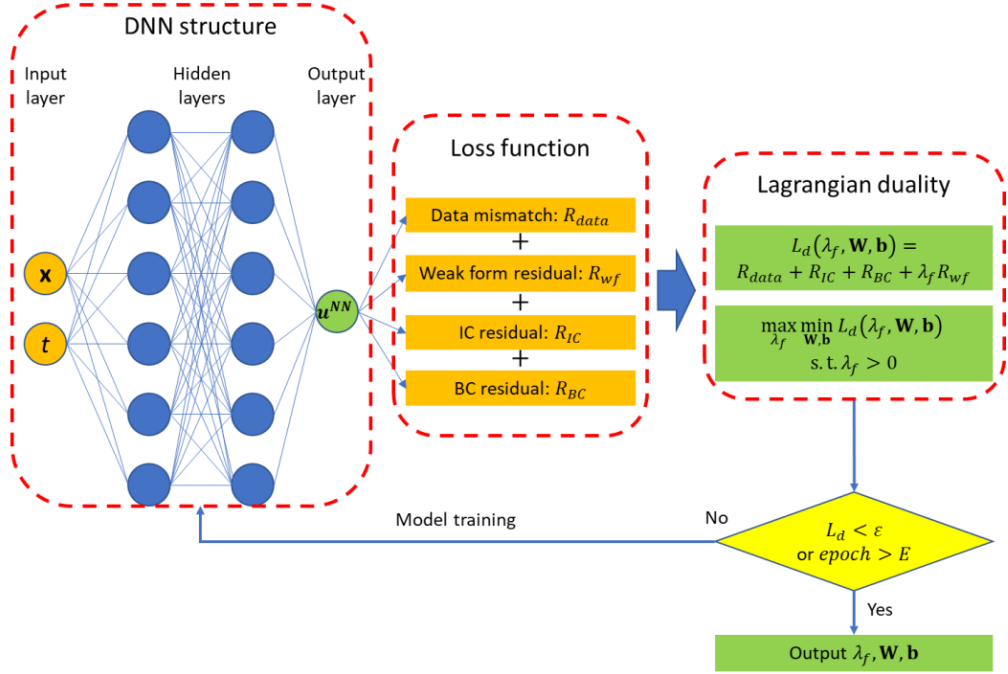

**Figure 1**. Schematic of the structure of the proposed TgNN-wf.

## 3 Case studies

In this section, we demonstrate the superiority of the proposed TgNN-wf over TgNN through two numerical examples that are commonly encountered in subsurface hydrological applications, including the hydraulic head prediction for the 2D unsteady-state single phase flow problem and the saturation profile prediction for the 1D unsteady-state two phase flow problem (Buckley-Leverett problem). For the first problem, we compare in detail the performance of TgNN-wf and TgNN in terms of accuracy and speed when different amounts of training data, collocation regions/points, or noises in the data are used. For the second problem in which discontinuity/shock in the saturation profile is expected, we compare the capability of TgNN-wf and TgNN to capture the shock when different amounts of data are used or different levels of noises are introduced.

3.1 2D unsteady-state single phase flow problem

We first consider a 2D unsteady-state single phase flow problem with a similar problem setup as in Wang et al. (2020). The governing equation of the problem can be written as:

$$S_s \frac{\partial h}{\partial t} = \frac{\partial}{\partial x}\left(K(x,y)\frac{\partial h}{\partial x}\right) + \frac{\partial}{\partial y}\left(K(x,y)\frac{\partial h}{\partial y}\right), \quad (13)$$

where $S_s = 0.0001$ is the specific storage, $h$ is the hydraulic head to be solved, and $K$ is the hydraulic conductivity field. We consider a 2D heterogeneous formation with size of 1020 [L]×1020 [L], where L is a consistent length unit. We use Karhunen–Loeve expansion (KLE) (Chang & Zhang, 2015; Zhang & Lu, 2004) with 20 terms to parameterize the random field $K$, assuming that $K$ obeys a lognormal distribution with $\langle \ln K \rangle = 0$, $\sigma_{\ln K} = 1$. **Fig. 2** shows a realization of $K$ used in the simulation. At $t = 0$, the left boundary is initialized with $h = 202$ [L], while the rest of the spatial domain is initialized uniformly with $h = 200$ [L]. The hydraulic heads at the left and right boundaries are kept constant to be 202 and 200 [L], respectively. The upper and lower boundaries are taken as no-flow boundaries. This problem can be solved numerically via the finite difference method by discretizing the spatial domain using a 51×51 grid. The numerical solution of the hydraulic head distribution for the first 50 timesteps with step size of 0.2 is obtained using the MODFLOW software (Harbaugh, 2005), which is then taken as the ground truth reference values. For illustration, as done in Wang et al. (2020), data from the first 18 timesteps are extracted as the training set, and we aim to train the network to predict the hydraulic head distribution of the remaining 32 timesteps.

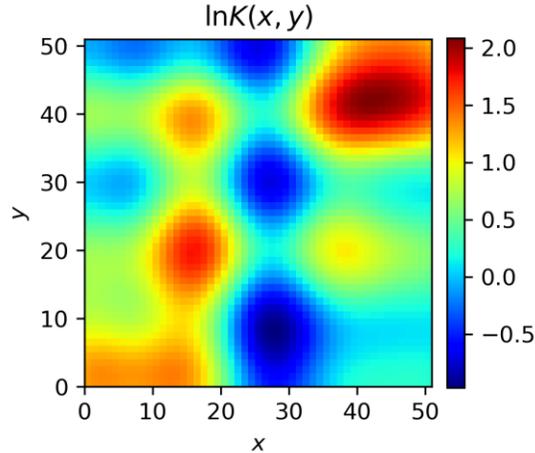

**Figure 2**. The hydraulic conductivity field $K$ used in the simulation. KLE with 20 terms was used

to parameterize the random field $K$ assuming a log normal distribution with $\langle \ln K \rangle = 0$, $\sigma_{\ln K} = 1$. Note that $x$ and $y$ coordinates are shown in grid numbers with grid spacing of 20.

The weak form formulation of Eq. 13 can be written as:

$$I = I_1 + I_2 + I_3 = 0 \tag{14.1}$$

$$I_1 = \int \omega S_s \frac{\partial h}{\partial t} dxdydt = -\int S_s h \frac{\partial \omega}{\partial t} dxdydt \tag{14.2}$$

$$I_2 = -\int \omega \frac{\partial}{\partial x}\left(K \frac{\partial h}{\partial x}\right) dxdydt = \int K \frac{\partial h}{\partial x} \frac{\partial \omega}{\partial x} dxdydt = -\int h \frac{\partial}{\partial x}\left(K \frac{\partial \omega}{\partial x}\right) dxdydt \tag{14.3}$$

$$I_3 = -\int \omega \frac{\partial}{\partial y}\left(K \frac{\partial h}{\partial y}\right) dxdydt = \int K \frac{\partial h}{\partial y} \frac{\partial \omega}{\partial y} dxdydt = -\int h \frac{\partial}{\partial y}\left(K \frac{\partial \omega}{\partial y}\right) dxdydt . \tag{14.4}$$

The three terms in Eq. 13 are integrated separately via $I_1$ to $I_3$. For Eq. 14.2-14.4 to hold true, we require test function $\omega$ and its first order derivatives with respect to $x$ and $y$ to vanish at the integration boundaries. A wide variety of test functions can be constructed accordingly, and here we define $\omega$ following the work by Reinbold et al. (2020):

$$\omega(x,y,t) = \left(\underline{x}^2 - 1\right)^2 \left(\underline{y}^2 - 1\right)^2 \left(\underline{t}^2 - 1\right) \tag{15.1}$$

$$\underline{x} = \frac{x - x_c}{H_x}, \quad \underline{y} = \frac{y - y_c}{H_y}, \quad \underline{t} = \frac{t - t_c}{H_t}, \tag{15.2}$$

where $(x_c, y_c, t_c)$ is the center coordinate of the integration subdomain, and $H_x$, $H_y$, and $H_t$ are the half length of the integral domain in $x$, $y$, and $t$ directions, respectively. Here we take $H_x = 0.1L_x$, $H_y = 0.1L_y$, and $H_t = 0.1L_t$, wherein $L_x$, $L_y$, and $L_t$ are the linear length of the simulation domain and $L_x = L_y = 1020$, $L_t = 10$. The effect of domain size on TgNN-wf performance will be shown later in Section 3.1.4.

To evaluate the integrals in Eq. 14.2-14.4, we use Gauss-Legendre quadrature rule (Kharazmi et al., 2019) with 5 integral points in each dimension (which results in 125 quadrature points for each subdomain). The FCNN used here consists of 7 hidden layers with 50 neurons in each layer, and we use softplus as the activation function (Zhao et al., 2018). As an illustration example, we randomly extracted 1,000 data points at each timestep from the first 18 timesteps as the training dataset (18,000 data points in total); 20,000 and 10,000 collocation points were used to impose BC and IC constraints, respectively. 10,000 collocation regions were sampled from the whole domain to impose the weak form constraint with their center coordinates determined using Latin Hypercube Sampling strategy (Raissi et al., 2019). At each iteration, Adam optimizer with learning rate 0.001 was used to minimize the problem $\min_{\mathbf{W},\mathbf{b}} L_d(\lambda_f, \mathbf{W}, \mathbf{b})$, then $\lambda_f$ is updated via maximizing Eq. 12 using gradient descent with a constant step size of 1.25. The model was trained in 1,000 epochs. The accuracy of the prediction results for the last 32 timesteps was evaluated using two metrics, including the relative $L_2$ error and the coefficient of determination ($R^2$ score), which are defined as:

$$L_2 = \frac{\|u^{NN} - u\|_2}{\|u\|_2}, \tag{16}$$

$$R^2 = 1 - \frac{\sum_{i=1}^{N_R}(u_i^{NN} - u_i)^2}{\sum_{i=1}^{N_R}(u_i - \bar{u})^2}, \tag{17}$$

where $\|\cdot\|_2$ represents the standard Euclidean norm, $N_R$ is the total number of evaluation points for $R^2$, and $\bar{u}$ is the average values of $u$.

**Fig. 3** shows the predicted hydraulic head distribution by TgNN-wf at the last timestep compared to the reference values obtained from the MODFLOW software. The predicted hydraulic heads along three horizontal lines ($y$=320, 620, and 920) are also shown and compared to the reference values. Good agreement between TgNN-wf predictions and reference values is observed, which can be further demonstrated in **Fig. 4**, where scattered points of predicted values vs. reference values closely align along the 45° straight line passing the origin. The resulting relative $L_2$ error is 7.02×10$^{-5}$ and $R^2$=0.9995. The optimized weight $\lambda_f$ for $R_{wf}$ in Eq. 9 is 13.5 (similarly, we get $\lambda_f$=90.9 for $R_{sf}$ in Eq. 5).

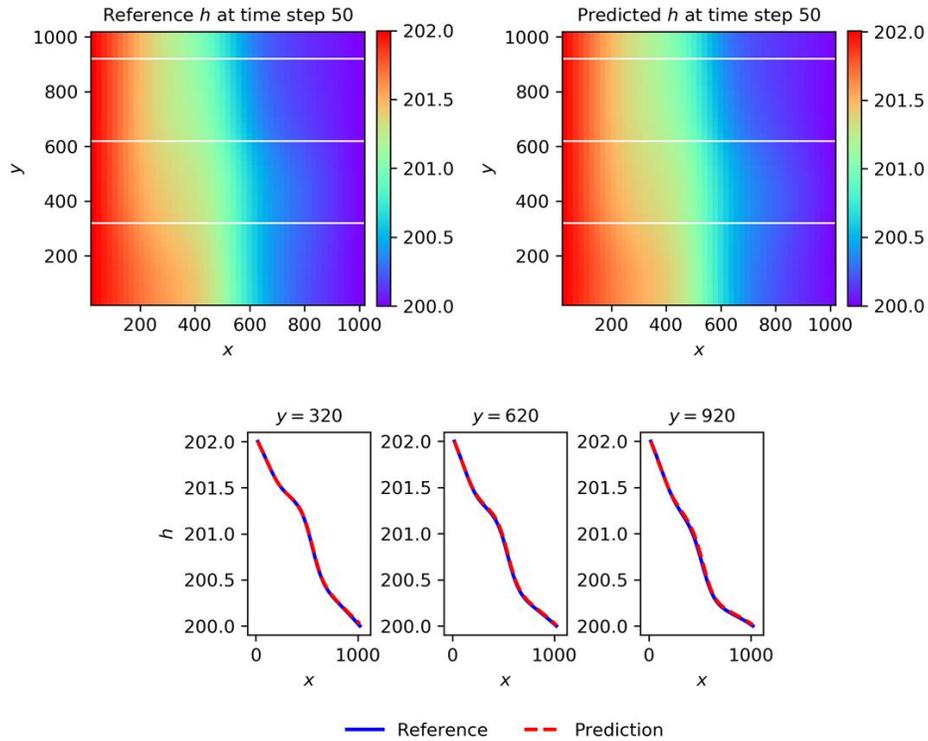

**Figure 3**. Hydraulic heads prediction at timestep 50 using TgNN-wf compared to reference values obtained by MODFLOW software. The upper panel shows the spatial hydraulic head distribution for reference values (left) and TgNN-wf predictions (right). The lower panel shows the hydraulic heads along three horizontal lines (drawn in white in the upper panel figures) *y*=320, 620, and 920 for reference values (solid curves) and TgNN-wf predictions (dashed curves).

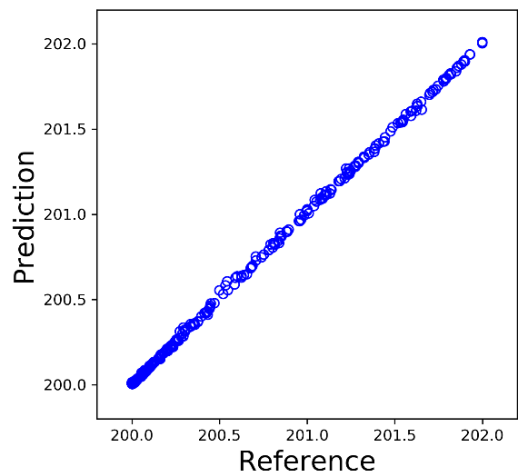

**Figure 4**. Correlation between TgNN-wf predictions and the reference values. Shown here are 300 randomly chosen data points from the results for the last 32 timesteps.

We then designed several case studies to compare the performance of TgNN-wf and TgNN in terms of computational efficiency, accuracy, and robustness to noises when different amount of training data or collocation points/regions are used.

3.1.1 Training with different amounts of data

We first tested the performance of TgNN and TgNN-wf to predict the hydraulic head distribution for the last 32 timesteps using different amounts of training data. For either model, 10, 100, and 1,000 data points randomly sampled from each timestep of the first 18 timesteps were used as the training data set, which results in 180, 1,800, and 18,000 training data points in total, respectively. We also considered a special case where no data constraint is imposed (0 training data). Note that we consistently used 1,000 collocation regions/points for TgNN-wf and TgNN, and both networks were trained with 1,000 epochs using a NVIDIA Tesla T4 GPU. For comparison, DNN results are shown as well.

**Fig. 5** shows the correlation plots between model predictions and reference values for TgNN-wf, TgNN, and DNN using different amounts of training data in which the upper, middle, and lower panels show the results using 180, 1,800, and 18,000 training data, respectively. **Table 1** shows the $L_2$ error, $R^2$ score, and elapsed training time for the three models. When only 180 training data were used, DNN has the worst performance ($R^2$=0.7604) due to the lack of theoretical guidance, while TgNN-wf and TgNN both have much higher accuracy ($R^2$=0.9938 and 0.9926) with the $L_2$ error about one order of magnitude smaller than DNN. Increasing the amount of training data improves the performance for DNN significantly. However, DNN easily tends to overfit and lacks generalization accuracy. On the other hand, TgNN-wf and TgNN are not sensitive to the amount of data used, and all cases have high enough accuracy. In general, the increase in training data results in slightly higher accuracy for both TgNN-wf and TgNN. It should be mentioned that even without the data constraint, both networks could achieve relatively good prediction results (see **Table 1**). When the same amount of training data is used, the accuracy of TgNN-wf is consistently higher than that of TgNN (though not significantly higher since both networks achieves high accuracy). Since the high order derivatives in the weak form formulation are transferred to the test functions whose derivatives can be analytically derived, the training process of TgNN-wf requires less operation of automatic differentiation. Therefore, the computational efficiency of TgNN-wf is higher than that of TgNN as shown in **Table 1**.

Furthermore, the computational burden for incorporating theoretical guidance is not significantly higher than DNN. For such a case, the improvement of TgNN-wf over TgNN is, however, not that significant, as the latter is already good enough (Wang et al., 2020).

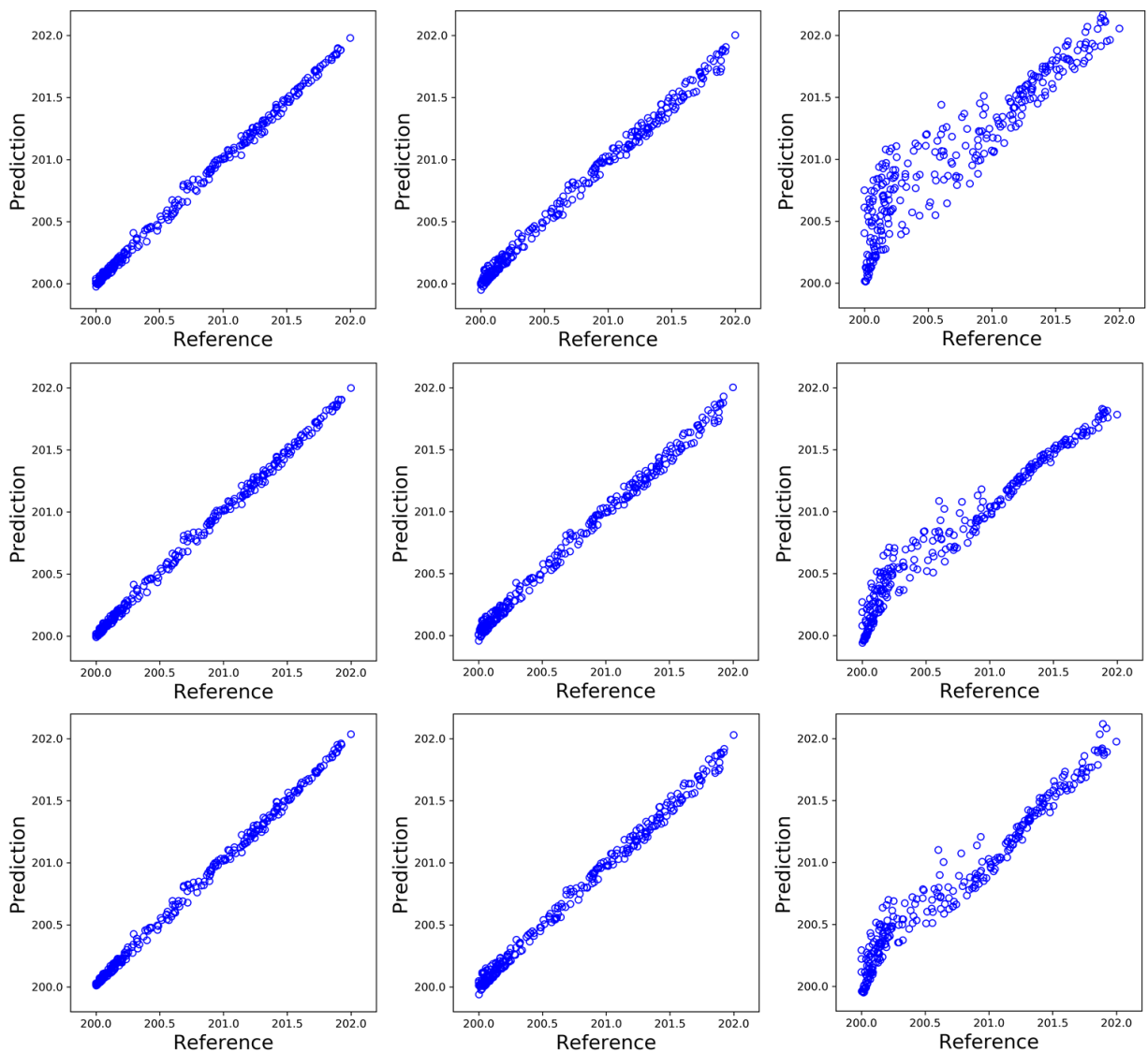

**Figure 5**. Correlation between model predictions and the reference values (left to right: TgNN-wf, TgNN, DNN). 300 randomly chosen prediction data points are shown in each subfigure. The upper, middle and lower panels show the results using 180, 1,800, and 18,000 training data points, respectively.

**Table 1.** Comparison of performance for TgNN-wf, TgNN, and DNN when different amounts of training data are used

| Training data points | TgNN-wf | | | TgNN | | | DNN | | |
|---|---|---|---|---|---|---|---|---|---|
| | $L_2$ error | $R^2$ | Training time, s | $L_2$ error | $R^2$ | Training time, s | $L_2$ error | $R^2$ | Training time, s |
| 0 | 3.21E-04 | 0.9898 | 77.00 | 3.44E-04 | 0.9882 | 87.63 | N/A | N/A | N/A |
| 180 | 2.50E-04 | 0.9938 | 77.88 | 2.72E-04 | 0.9926 | 88.44 | 1.55E-03 | 0.7604 | 38.18 |
| 1800 | 2.28E-04 | 0.9948 | 91.59 | 2.49E-04 | 0.9938 | 104.76 | 6.90E-04 | 0.9526 | 50.54 |
| 18000 | 2.12E-04 | 0.9955 | 241.88 | 2.27E-04 | 0.9949 | 247.93 | 6.39E-04 | 0.9593 | 203.82 |

3.1.2 Training with different amounts of collocation regions/points

**Table 2** shows the performance comparison between TgNN-wf and TgNN using different amounts of collocation regions/points. Here we kept the number of training data points to be constant of 1,800, and both networks were trained with 1,000 epochs. When the networks are trained only on data without informed by theoretical constraints (0 collocation points/regions), the performance is generally poor. With only 100 collocation regions/points, the performance is notably improved, reducing $L_2$ error approximately by half. As expected, the prediction accuracy of both networks increases with the amount of collocation regions/points used, while the performance of TgNN-wf is consistently better than that of TgNN. It should be mentioned that even though the amount of collocation regions/points is the same for TgNN-wf and TgNN, the total times of governing equation evaluation are significantly different. For TgNN-wf, the governing equation is evaluated 125 times on each collocation region, while the governing equation is evaluated only once for a collocation point in TgNN. However, TgNN-wf trains faster than TgNN unless a large amount of collocation regions (e.g., 10,000) is used when the time needed for governing equation evaluation is comparable to automatic differentiation and can no longer be ignored. In practice, a much smaller amount of collocation regions is needed for TgNN-wf to achieve similar prediction accuracy to TgNN.

**Table 2**. Comparison of performance between TgNN-wf and TgNN when different amounts of collocation regions/points are used

| collocation regions/points | TgNN-wf | | | TgNN | | |
|---|---|---|---|---|---|---|
| | $L_2$ error | $R^2$ | Training time, s | $L_2$ error | $R^2$ | Training time, s |
| 0 | 6.90E-04 | 0.9526 | 50.54 | 6.90E-04 | 0.9526 | 50.54 |
| 100 | 3.79E-04 | 0.9857 | 60.04 | 4.23E-04 | 0.9822 | 73.40 |
| 1,000 | 2.28E-04 | 0.9948 | 91.59 | 2.49E-04 | 0.9938 | 104.76 |
| 10,000 | 2.25E-04 | 0.9950 | 406.49 | 2.43E-04 | 0.9941 | 130.22 |

3.1.3 Training with noisy data

We then compared the performance of TgNN and TgNN-wf in the presence of noisy data. Here the noises are introduced in the following manner:

$$h^*(x, y, t) = h(x, y, t) + \Delta h(x, y)a\varepsilon, \tag{18}$$

where $h$ and $h^*$ are the clean and noisy data, respectively; $\Delta h$ is the maximum difference of the observed hydraulic head at location $(x, y)$ during the first 18 timesteps; $a$ is the percentage of noises to be added; and $\varepsilon$ is a random variable ranging from -1 to 1. We kept the number of training data to be constant of 1800 and tested the performance of TgNN-wf and TgNN with 10%, 20%, 40%, 60%, and 80% noises added to the training data. The results are shown in **Table 3** with DNN results shown as well for comparison. In general, DNN has the worst performance and its accuracy is dramatically affected by the amount of noises introduced, with $L_2$ error about an order of magnitude larger than TgNN-wf. The accuracy of TgNN-wf, however, is not obviously undermined even with 80% noise and is consistently the highest of the three models. TgNN is more sensitive to noises in the training data than TgNN-wf, and its accuracy decreases with the amount of noises introduced. This can be explained by the integration nature of TgNN-wf, in which the noises in the training data are smoothed or averaged by evaluating the weak form integrals in each subdomain.

**Table 3**. Comparison of performance between TgNN-wf and TgNN when different amounts of noises are introduced into the training dataset

| noise, % | vTgNN | | TgNN | | DNN | |
|---|---|---|---|---|---|---|
| | $L_2$ error | $R^2$ | $L_2$ error | $R^2$ | $L_2$ error | $R^2$ |
| 0 | 2.28E-04 | 0.9948 | 2.49E-04 | 0.9938 | 6.90E-04 | 0.9526 |
| 10 | 2.24E-04 | 0.9950 | 2.76E-04 | 0.9924 | 1.02E-03 | 0.8973 |
| 20 | 2.30E-04 | 0.9947 | 3.72E-04 | 0.9862 | 1.36E-03 | 0.8157 |
| 40 | 2.34E-04 | 0.9946 | 4.33E-04 | 0.9814 | 1.43E-03 | 0.7966 |
| 60 | 2.38E-04 | 0.9944 | 4.54E-04 | 0.9795 | 1.54E-03 | 0.7631 |
| 80 | 2.49E-04 | 0.9938 | 4.86E-04 | 0.9765 | 1.62E-03 | 0.7387 |

3.1.4 Training with collocation regions of different sizes for TgNN-wf

We then studied the effect of the size of the collocation region on the prediction performance of TgNN-wf. Three different cases were tested with relative collocation region size of 0.08, 0.2, and 0.4, and the relative size is defined as the fraction of the linear length of the subdomain to the whole domain. The corresponding number of Gaussian quadrature points is determined proportional to the subdomain size. We kept the number of training data points, collocation regions, and training epochs to be 1,800, 500, and 1,000, respectively. The prediction accuracy is shown in **Table 4**. For this specific problem, we observe that decreasing the size of the collocation region improves the prediction accuracy, due to local fluctuations in the solution that can be better captured by evaluating the integrals on smaller subdomains. It is worth mentioning that using a smaller collocation region also decreases the computational cost since fewer Gaussian quadrature points are needed for integral evaluation.

**Table 4**. Comparison of TgNN-wf performance using different collocation region sizes

| # of Gaussian quadrature points | Relative size of the collocation region | # of evaluation points | $L_2$ error | $R^2$ |
|---|---|---|---|---|
| 64 | 0.08 | 32,000 | 2.05E-04 | 0.9958 |
| 1,000 | 0.2 | 500,000 | 2.28E-04 | 0.9948 |
| 8,000 | 0.4 | 4,000,000 | 2.62E-04 | 0.9932 |

### 3.2 1D unsteady-state Buckley-Leverett two phase flow problem

In Section 3.1, we dealt with a relatively simple hydrology problem where only one phase is involved in the flow process and the solution is relatively smooth. In this section, we test the robustness of our proposed TgNN-wf on a more challenging problem: where two-phase flow is considered and a sharp discontinuity is expected in the solution space, which will be problematic for TgNN since the high-order derivative evaluation around the discontinuity causes numerical instability.

We here consider a situation where water is displacing oil in a horizontal reservoir. The problem can be simplified in a 1D setup along the flow streamline, and we neglect the effect of gravity or capillarity. The unsteady water saturation evolvement along the displacement pathway can be described by the Buckley-Leverett equation (Buckley & Leverett, 1942; Zhang et al., 2000):

$$\frac{\partial S(\tau,t)}{\partial t} + f_w'(S)\frac{\partial S(\tau,t)}{\partial \tau} = 0, \tag{19}$$

where $S$ is the water saturation; $f_w(S)$ is the fractional flow of water which is a function of $S$; $\tau$ is the travel time of a particle from $x_0$ to $x$ in the velocity field $u(x)$; and $d\tau/dx = 1/u(x)$. Consider the case that the formation is initially saturated with oil with $S(\tau,0) = S_{wr}$ and a continuous water injection at rate $q$ is placed at $\tau = 0$ and $S(0,t) = 1 - S_{or}$, where $S_{wr}$ and $S_{or}$ are the residual water and oil saturations respectively. The analytical solution of Eq. 19 can be written as (Zhang & Tchelepi, 1999):

$$S(\tau,t) = s\left(\frac{\tau}{t}\right) H\left[f_w'(S_*) - \frac{\tau}{t}\right], \tag{20}$$

where $S_*$ is the solution of $f_w'(S) = [f_w(S) - f_w(S_{wr})]/(S - S_{wr})$. $S_*$ is the saturation at the shock front, where saturation drops from $S_*$ behind the shock to $S_{wr}$ in front of the shock, forming a discontinuity. $s(\tau/t)$ is the solution of $f_w'(S) = \tau/t$, and $H$ is the Heaviside step function.

For TgNN, neural network approximation of the squared residual based on Eq. 19 is incorporated into the loss function (Eq. 5). For TgNN-wf, the weak form formulation of this problem can be written as:

$$I = \int_{\Delta x \times \Delta t}\left(\omega\frac{\partial S}{\partial t} + \omega\frac{\partial f_w}{\partial \tau}\right)d\tau dt = \int_{\Delta x \times \Delta t} -\left(S\frac{\partial \omega}{\partial t} + f_w\frac{\partial \omega}{\partial \tau}\right)d\tau dt, \tag{21}$$

subject to the condition that $\omega$ vanishes at the integration boundaries. Here we define $\omega$ as:

$$\omega(\tau, t) = (\underline{\tau}^2 - 1)(\underline{t}^2 - 1), \tag{22}$$

where $\underline{\tau} = (\tau - \tau_c)/H_\tau$, $\underline{t} = (t - t_c)/H_t$. $(\tau_c, t_c)$ is the center coordinate of the integration subdomain, and $H_\tau$ and $H_t$ are the half length of the subdomain in $\tau$ and $t$ directions, respectively.

We set the water injection rate $q = 0.3$, the reservoir porosity $\phi = 0.3$, and we assume the porosity is constant along the flow direction. Therefore, the total flux $u$ is constant 1. For simplicity, we consider the following Corey-type relative permeability curves with $S_{wr} = S_{or} = 0$,

$$k_{rw} = S^2, k_{ro} = (1-S)^2. \tag{23}$$

Therefore,

$$f_w(S) = \frac{mS^2}{mS^2 + (1-S)^2}, \tag{24}$$

$$f_w'(S) = \frac{2mS(1-S)}{[mS^2 + (1-S)^2]^2}, \tag{25}$$

where $m$ is the viscosity ratio and here we set $m = 2$.

For both TgNN-wf and TgNN, we used a network structure of 8 hidden layers with 20 neurons per layer. Tanh (Goodfellow et al., 2016) was used as the activation function, and we used Adam optimizer with learning rate of 0.001 to train both networks with 10,000 epochs. In the model training process, 2,000 collocation points were randomly selected to impose IC and BC constraints. 10,000 collocation regions/points were extracted to impose strong or weak form constraints based on Latin hypercube sampling strategy. For TgNN-wf with the collocation region size $H_\tau = H_t = 0.004$, 10 Gauss-Legendre quadrature points were used for numerical integration along each dimension. The analytical solution was calculated on 501 uniformly spaced points on x axis ($x \in [0, 1]$) with spacing of 0.002. 50 timesteps were calculated ($t \in (0,1)$) with spacing of 0.02. The training data were obtained from the first 20 timesteps, and we use both models to predict the saturation profile for the remaining 30 timesteps. In the following sections, we compare TgNN-wf and TgNN when different amounts of training data are used or different levels of noises in the training data are introduced.

### 3.2.1 Training with different amounts of data

We first tested the performance of TgNN-wf and TgNN using different amounts of data. For either model, 10, 100, and 500 data points randomly sampled from each timestep of the first 20 timesteps were used as the training data set, which results in 200, 2,000, and 10,000 training data points in total, respectively. We also considered a special case where no data constraint was imposed (0 training data). For comparison, DNN results were shown as well. **Figs. 6, 7,** and **8** shows the predicted saturation distributions of DNN, TgNN, and TgNN-wf at four timesteps in comparison with the reference values solved analytically by Eq. 20. **Table 5** shows the prediction accuracy for the three models. When only 200 data points are used, DNN incorrectly predicts the shock location for $t$=0.4 and 0.6, and falsely predicts the presence of a shock after water breakthrough ($t$=0.8). Though increasing the amount of training data improves the prediction accuracy for DNN, the results are not satisfactory for later timesteps (e.g., $t$=0.6 and 0.8).

Due to the complexity of the problem, neither TgNN-wf or TgNN could accurately solve the PDE without labelled data. TgNN failed to capture the occurrence of shock at all, and it only provides satisfactory results behind the shock, due to the inaccurate approximation of the derivatives around the shock. On the other hand, TgNN-wf accurately captures the shock location and provides satisfactory results in front of the shock, resulting in the highest accuracy of the three models. However, significant overshooting is observed behind the shock. In general, both models perform relatively well after water breakthrough. However, the results are far from satisfactory before water breakthrough as a result of the strong discontinuity in the solution space. Increasing the amount of training data helps TgNN locate the shock position. However, significant numerical diffusion is observed near the shock that noticeably smoothes the shock, which is a common characteristic for numerical methods as well. Further increasing the training data does not solve this problem, and the overall accuracy for TgNN is lower than DNN. This is because the performance of TgNN is limited by the incorporation of a strong form constraint, which inevitably introduces numerical diffusion near the shock. On the other hand, the overshooting problem for TgNN-wf is mitigated using more training data and the accuracy is the highest of the three models. The shock position is captured accurately and the sharpness of the shock is maintained. This demonstrates the superiority of weak form over strong form for solving discontinuous problems. Nevertheless, there is still noticeable overshooting or undershooting behavior near the shock for TgNN-wf.

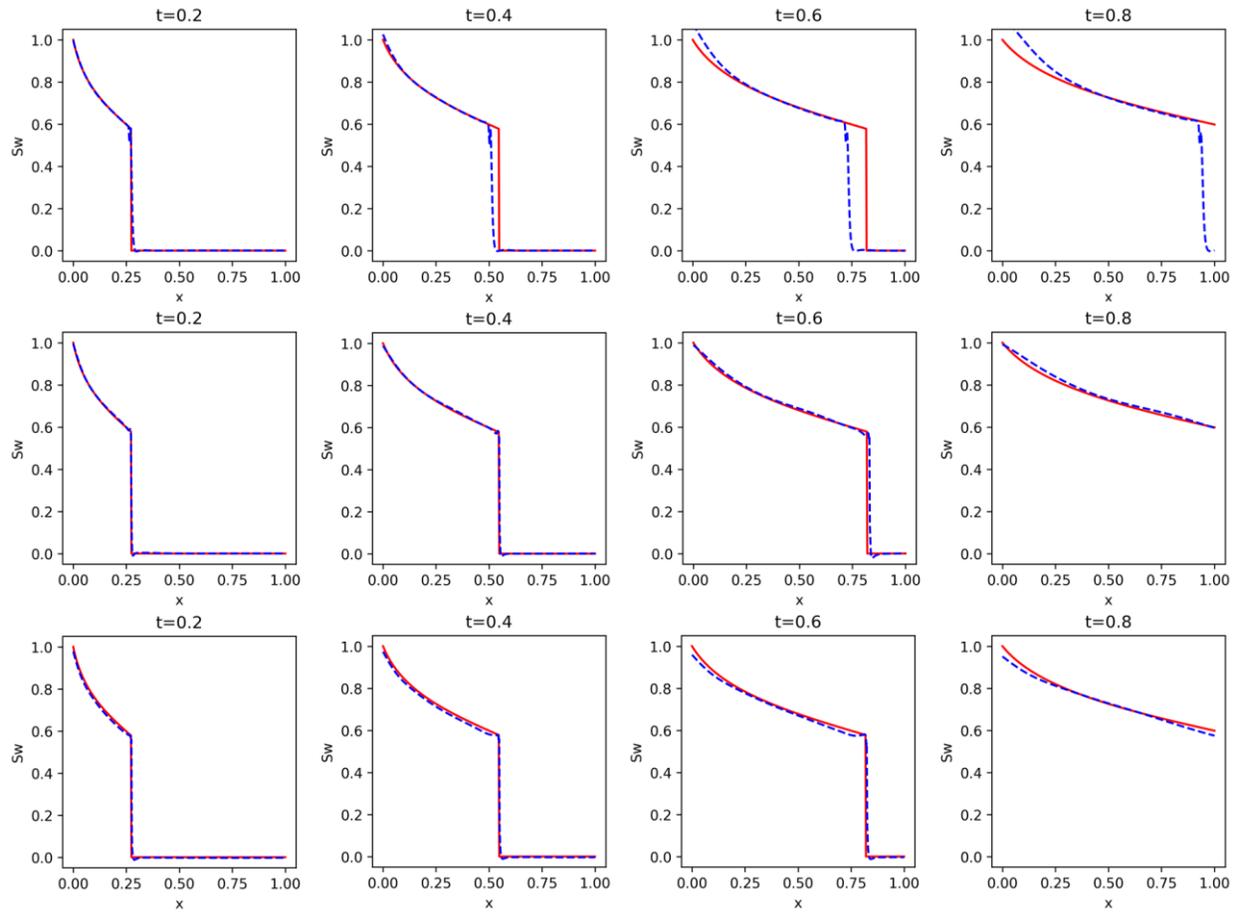

**Figure 6**. Predicted saturation profiles of DNN at four timesteps. The upper, middle, and lower panels show the results using 200, 2,000, and 10,000 training data points, respectively. The dashed curves are prediction results and the solid curves are analytical solutions.

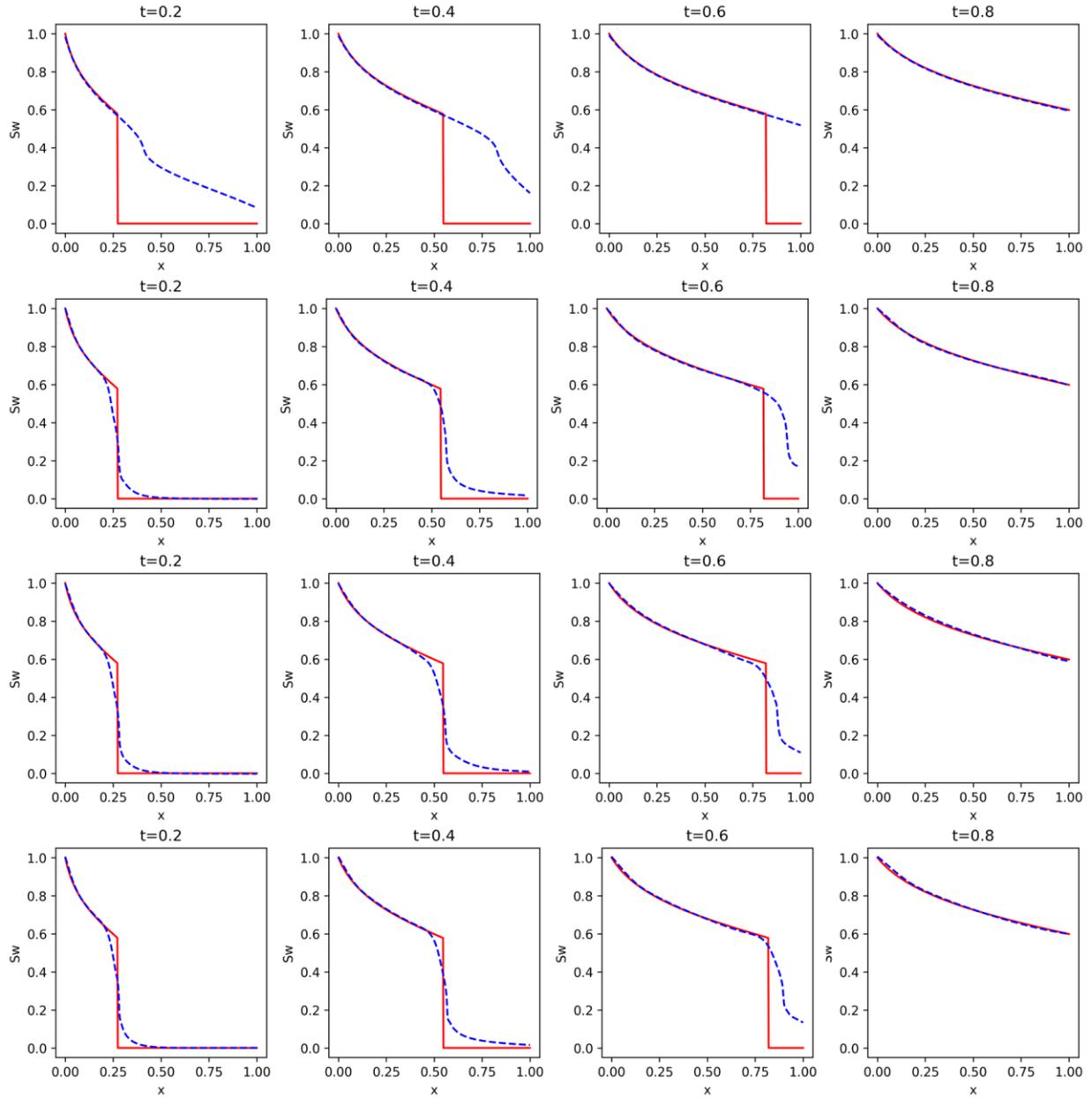

**Figure 7**. Predicted saturation profiles of TgNN at four timesteps. The panels from top to bottom show the results using 0, 200, 2,000, and 10,000 training data points, respectively. The dashed curves are prediction results and the solid curves are analytical solutions.

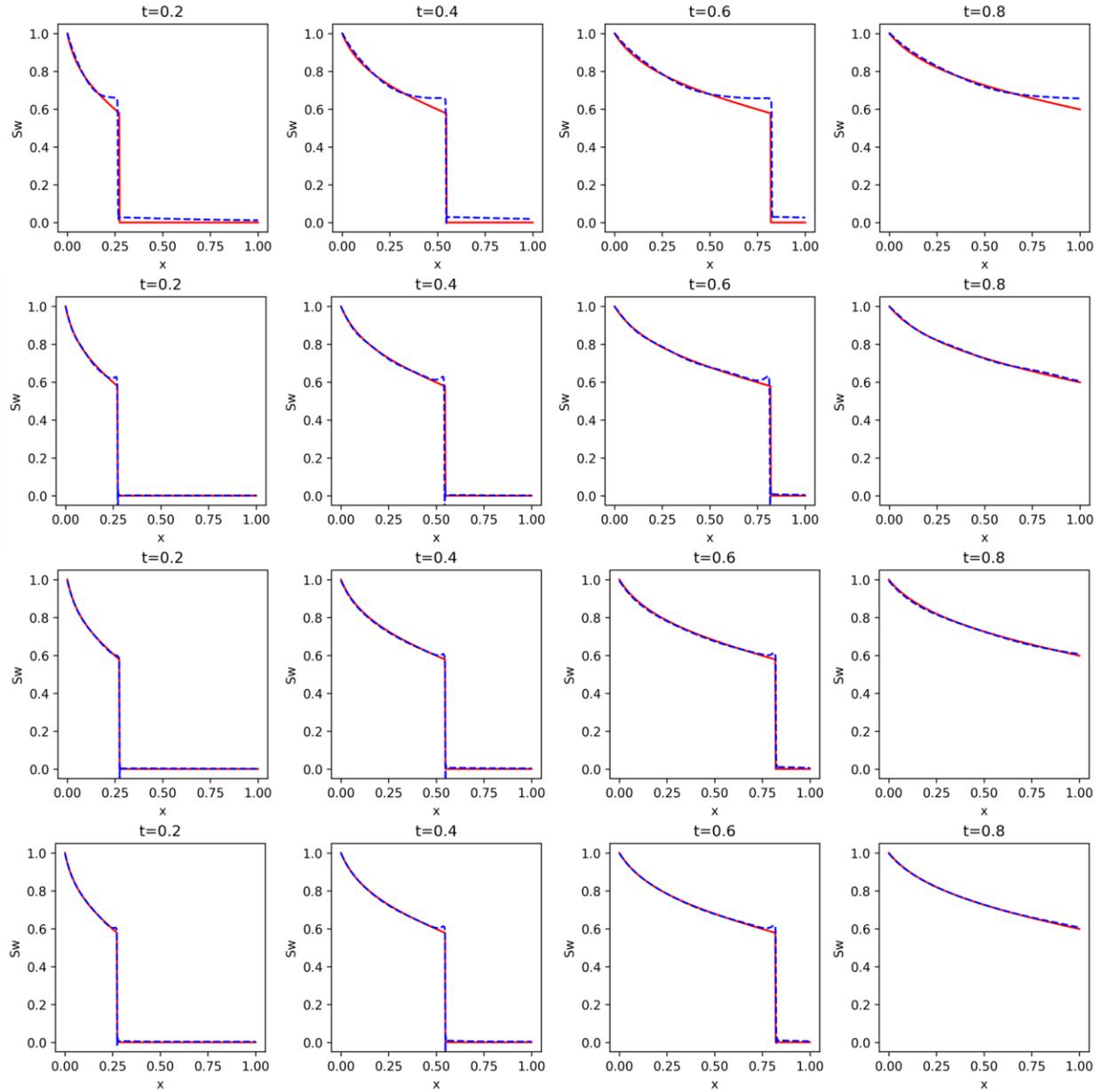

**Figure 8**. Predicted saturation profiles of TgNN-wf at four timesteps. The panels from top to bottom show the results using 0, 200, 2,000, and 10,000 training data points, respectively. The dashed curves are prediction results and the solid curves are analytical solutions.

**Table 5**. Comparison of performance for TgNN-wf, TgNN, and DNN using different amounts of training data

| Training data points | TgNN-wf | | TgNN | | TgNN-wf with diffusion term | | TgNN with diffusion term | | DNN | |
|---|---|---|---|---|---|---|---|---|---|---|
| | $L_2$ error | $R^2$ | $L_2$ error | $R^2$ | $L_2$ error | $R^2$ | $L_2$ error | $R^2$ | $L_2$ error | $R^2$ |
| 0 | 7.24E-02 | 0.9858 | 3.48E-01 | 0.6728 | 6.20E-02 | 0.9896 | 2.43E-01 | 0.8400 | N/A | N/A |
| 200 | 4.27E-02 | 0.9951 | 1.49E-01 | 0.9401 | 4.96E-02 | 0.9934 | 5.88E-02 | 0.9906 | 1.89E-01 | 0.9035 |
| 2,000 | 2.36E-02 | 0.9985 | 1.10E-01 | 0.9674 | 2.94E-02 | 0.9977 | 5.69E-02 | 0.9912 | 5.83E-02 | 0.9908 |
| 10,000 | 2.42E-02 | 0.9984 | 1.20E-01 | 0.9612 | 1.62E-02 | 0.9993 | 2.93E-02 | 0.9977 | 4.81E-02 | 0.9937 |

3.2.2 Training with different amounts of data with the added diffusion term

Here we attempt to smooth the TgNN-wf predictions by artificially introducing a diffusion term to the right-hand side of the governing equation (Eq. 19), and we get:

$$\frac{\partial S(\tau,t)}{\partial t} + f'_w(S)\frac{\partial S(\tau,t)}{\partial \tau} = \eta \frac{\partial^2 S(\tau,t)}{\partial \tau^2}, \quad (26)$$

where $\eta$ is the diffusion coefficient. This treatment has also been done before in numerical solutions of the Buckley-Leverett equation (Aziz & Settari, 1979; Langtangen et al., 1992). If $\eta$ is small enough, the shock can be roughly described by the continuous solution of Eq. 26. The weak form formulation of Eq. 26 can be written as:

$$I = \int_{\Delta x \times \Delta t} \left( \omega \frac{\partial S}{\partial t} + \omega \frac{\partial f_w}{\partial \tau} - \omega \eta \frac{\partial^2 S}{\partial \tau^2} \right) d\tau dt = \int_{\Delta x \times \Delta t} -\left( S \frac{\partial \omega}{\partial t} + f_w \frac{\partial \omega}{\partial \tau} + S\eta \frac{\partial^2 \omega}{\partial \tau^2} \right) d\tau dt, \quad (27)$$

which requires that $\omega$ vanishes at $t$ and $\tau$ boundaries and $\partial \omega / \partial \tau$ vanishes at $\tau$ boundaries. Here we define $\omega$ as:

$$\omega(\tau, t) = (\underline{\tau}^2 - 1)^2 (\underline{t}^2 - 1). \quad (28)$$

The theoretical guidance for TgNN-wf and TgNN is constructed using Eq. 27 and 26, respectively, and similarly we tested the performance of TgNN-wf and TgNN to solve the PDE using different amounts of data. **Figs. 9** and **10** show the predicted saturation distributions of TgNN and TgNN-wf at four timesteps in comparison with the reference values using 0, 200, 2,000, and 10,000 training data. The error quantification results are shown in **Table 5**. By introducing the diffusion term, the prediction accuracy of both models is dramatically improved. The solution of TgNN shows the presence of a shock without labelled data for the first time, though the predicted shock location is consistently ahead of the true location. Increasing the amount of training data helps TgNN locate the shock position. The sharpness of the shock is captured more accurately than former cases without the diffusion term, which indicates the effectivity of introducing the small diffusion term in solving discontinuous problems. For TgNN-wf, the prediction accuracy without labelled data is higher than the case without the diffusion term (see **Table 5**). The overshooting problem near the shock is resolved by adding the diffusion term. Increasing the amount of training data further increases the prediction accuracy for TgNN-wf, and the overshooting and undershooting problems near the shock are resolved while maintaining the sharpness of the shock. It should be noted that the predicted results for TgNN still experience larger numerical diffusion near the shock than TgNN-wf. We have tested different values of the diffusion coefficient for both models: a larger coefficient leads to more significant numerical diffusion near the shock. However, when the coefficient is too small, the results are similar to the cases without the diffusion term. The results shown here are obtained using diffusion coefficients of $2.0 \times 10^{-3}$ and $4.0 \times 10^{-4}$ for TgNN and TgNN-wf, respectively, which are the minimal values achievable before the prediction accuracy near the shock begins to deteriorate. By introducing the weak form formulation of the problem into the loss function, we could use a diffusion coefficient about one order of magnitude smaller than TgNN to better capture the sharpness of the shock and reduce the effect of numerical diffusion. This further demonstrates the robustness of TgNN-wf to deal with strong discontinuity in the solution space. Moreover, the performance of TgNN-wf in the small data regime (<200) is much better than TgNN or DNN.

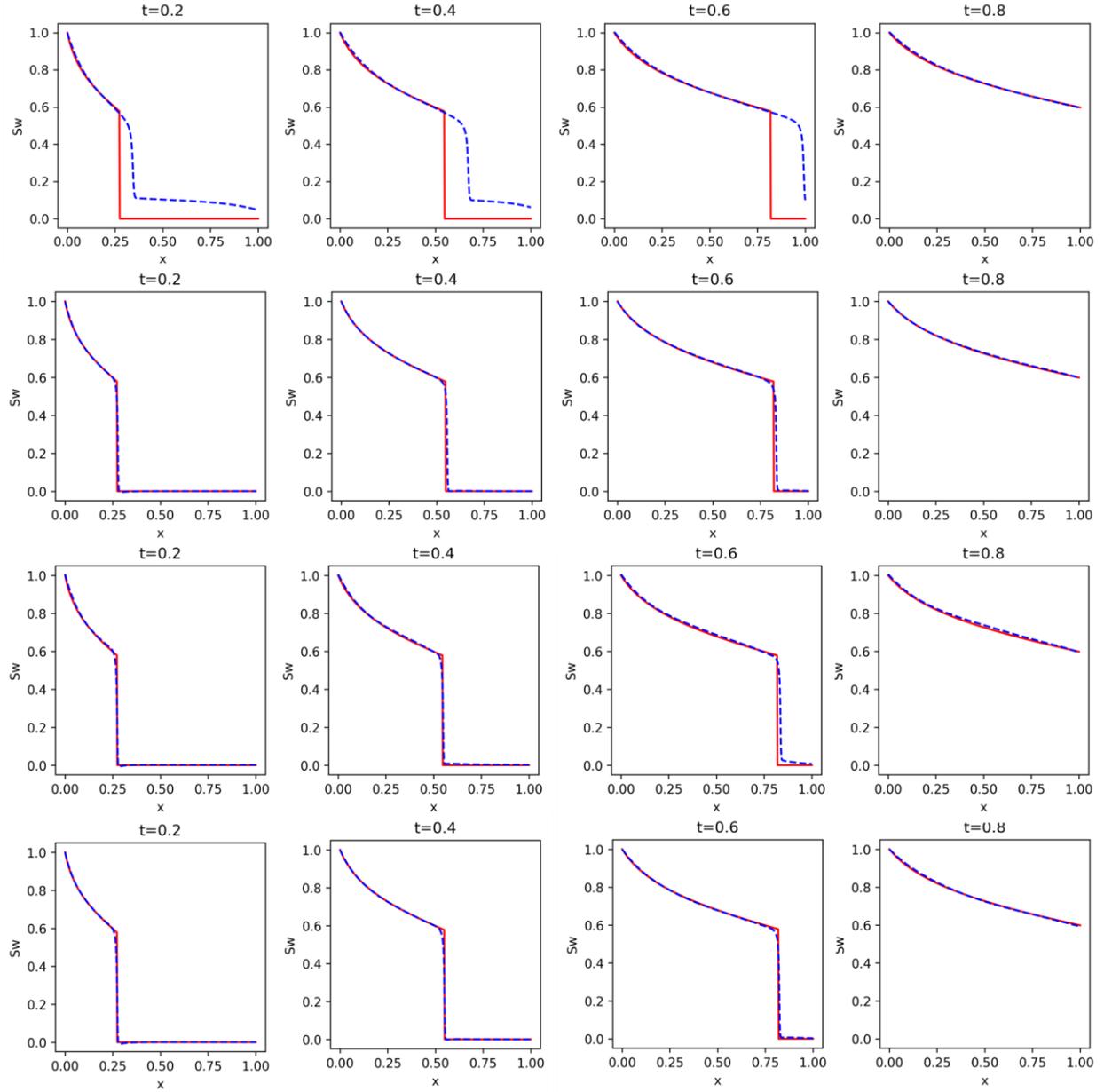

**Figure 9**. Predicted saturation profiles of TgNN with the diffusion term at four timesteps. The panels from top to bottom show the results using 0, 200, 2,000, and 10,000 training data points, respectively. The dashed curves are prediction results and the solid curves are analytical solutions.

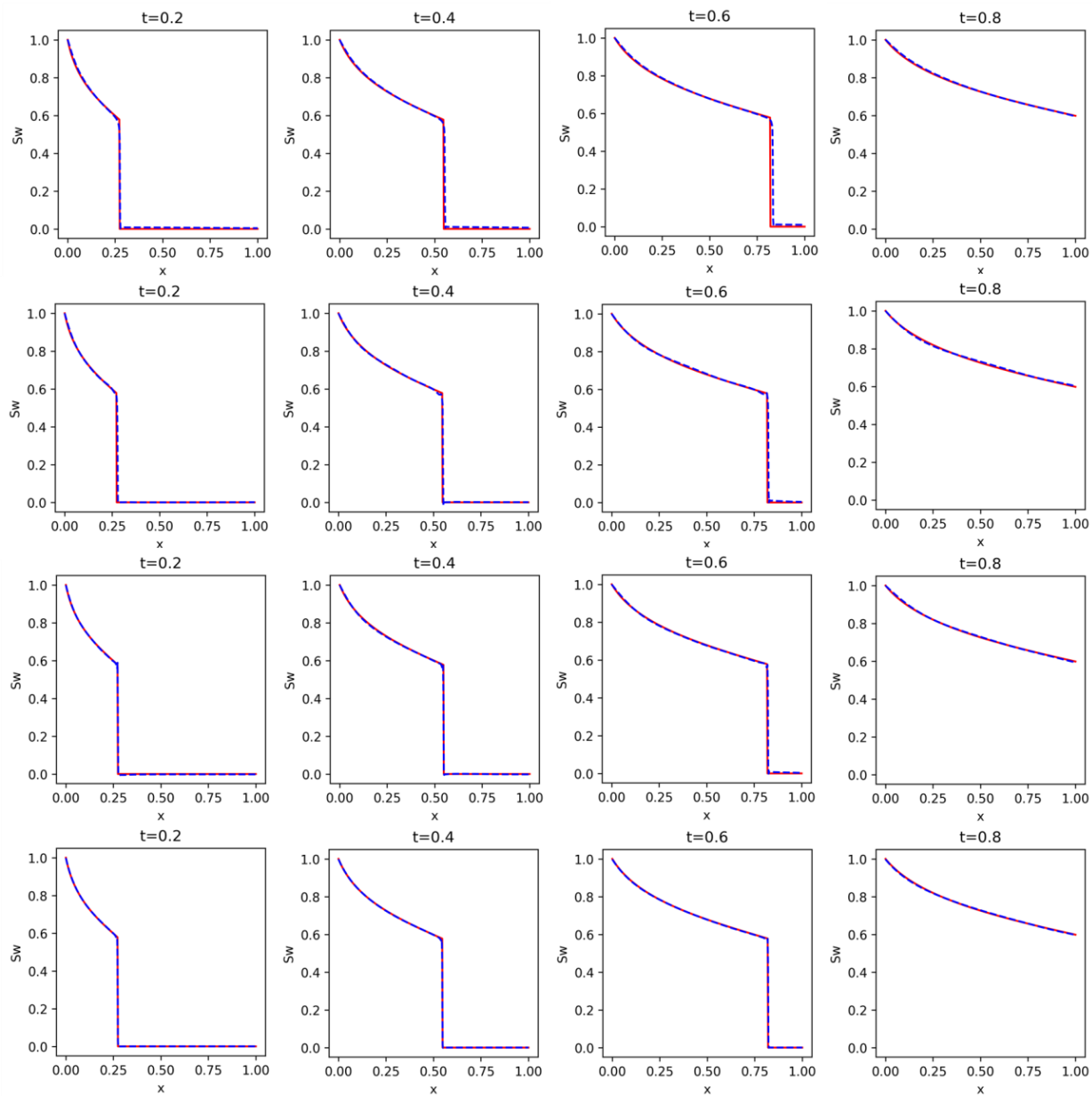

**Figure 10**. Predicted saturation profiles of TgNN-wf with the diffusion term at four timesteps. The panels from top to bottom show the results using 0, 200, 2,000, and 10,000 training data points, respectively. The dashed curves are prediction results and the solid curves are analytical solutions.

3.2.3 Training with noisy data with the added diffusion term

Finally, we tested the performance of TgNN-wf and TgNN when different levels of noises are introduced into the training data. 10%, 20%, 40%, 60%, and 80% noises were introduced to the 200 training data in the same way as described in Section 3.1.3, and the results are shown in **Table 6**. The prediction accuracy of DNN is the lowest of the three models and decreases dramatically with the amount of noises introduced. For noises higher than 40%, no meaningful prediction can be made which results in negative $R^2$ scores, therefore the results are not shown in **Table 6**. The performance of TgNN-wf, however, is not obviously affected by the quality of the training data up to 20% noises, after which the accuracy slowly decreases with the amount of noises introduced. Even with 80% noises in the training data, TgNN-wf achieves acceptable accuracy with $L_2$ error only about 50% higher than the case without noises. On the other hand, the accuracy of TgNN consistently decreases with the amount of noises introduced, and the $L_2$ error for the case with 80% noises is about four times larger than the case without noises. Again, this demonstrates the superiority of TgNN-wf over TgNN in terms of robustness to noises.

**Table 6**. Comparison of performance for TgNN-wf and TgNN with the diffusion term and DNN using 200 training data points with different amount of noises

| noise, % | TgNN-wf with diffusion term | | TgNN with diffusion term | | DNN | |
|---|---|---|---|---|---|---|
| | $L_2$ error | $R^2$ | $L_2$ error | $R^2$ | $L_2$ error | $R^2$ |
| 0 | 4.96E-02 | 0.9934 | 5.88E-02 | 0.9906 | 1.89E-01 | 0.9035 |
| 10 | 5.43E-02 | 0.9920 | 6.23E-02 | 0.9895 | 2.34E-01 | 0.8515 |
| 20 | 5.41E-02 | 0.9921 | 1.32E-01 | 0.9531 | 2.50E-01 | 0.8307 |
| 40 | 6.35E-02 | 0.9891 | 1.54E-01 | 0.9361 | × | × |
| 60 | 7.09E-02 | 0.9864 | 1.97E-01 | 0.8945 | × | × |
| 80 | 7.29E-02 | 0.9856 | 2.11E-01 | 0.8790 | × | × |

## 4 Conclusions

We propose a TgNN-wf framework, which incorporates the weak form formulation of the physical problem to construct theoretical guidance for the deep neural network. By performing integration-by-parts, high-order derivatives in the original PDE can be effectively transferred to the test functions, which can be artificially defined to reduce the computational cost. Domain decomposition was used to reduce the number of Gaussian quadrature points needed for evaluating the integrals, with the potential to incorporate locally defined test functions and quadrature rules to capture the local discontinuity or fluctuation. Two numerical cases commonly encountered in the subsurface flow problems were studied, including the unsteady-state 2D single phase flow problem and the unsteady-state 1D two-phase flow problem (Buckely-Leverett problem). We systematically compared the performance of TgNN and TgNN-wf in terms of accuracy, robustness to noises, and computational efficiency. We found that TgNN-wf results in higher accuracy, especially for the Buckley-Leverett problem when strong discontinuity in the solution space is present. TgNN-wf not only captures the shock position, but also maintains the sharpness of the shock, while noticeable numerical diffusion effect is observed near the shock for TgNN predictions. TgNN-wf maintains the highest prediction accuracy in the small data regime compared to TgNN and DNN. Moreover, the performance of TgNN-wf is not obviously deteriorated with noisy training data (with noises up to 80% or 20% for the single-phase and two-phase flow cases, respectively), while TgNN is more sensitive to noises. The computational efficiency of TgNN-wf is higher than that of TgNN, unless a large number of collocation regions/points are used, which is most likely not necessary since TgNN-wf could achieve comparable accuracy with a smaller amount of collocation regions.

This work paves the way for which a large variety of complex problems (e.g. small or noisy training data, high dimensional inputs, strong discontinuity or fluctuation) can be solved via theory-guided deep learning more accurately and efficiently. Here we fixed the test functions to be polynomials and used one test function for each integral subdomain. Different forms of test functions (e.g. sin functions or Gauss-Legendre polynomials) have been proposed in the literature, and we will compare the performance of different test functions in future work. In the Lagrangian duality formulation of the loss function, we focused on optimizing the weight of the weak form residual term. A more rigorous study to optimize the weights of all the terms is deferred to future work.

**Acknowledgments and Data**

We acknowledge Pengcheng Cloud Brain at Peng Cheng Laboratory for providing high performance GPU computing resources. The supporting data can be found in Xu (2020).

**References**

Almasri, M. N., & Kaluarachchi, J. J. (2005). Groundwater Flow and Transport Process. In *Water*

*Encyclopedia* (pp. 514–518). American Cancer Society. https://doi.org/10.1002/047147844X.gw411

Aziz, K., & Settari, A. (1979). *Petroleum Reservoir Simulation*. Springer Netherlands.

Bao, G., Ye, X., Zang, Y., & Zhou, H. (2020). Numerical Solution of Inverse Problems by Weak Adversarial Networks. *ArXiv:2002.11340 [Cs, Math]*. http://arxiv.org/abs/2002.11340

Blunt, M. J. (2001). Flow in porous media—Pore-network models and multiphase flow. *Current Opinion in Colloid & Interface Science*, *6*(3), 197–207. https://doi.org/10.1016/S1359-0294(01)00084-X

Buckley, S. E., & Leverett, M. C. (1942). Mechanism of Fluid Displacement in Sands. *Transactions of the AIME*, *146*(01), 107–116. https://doi.org/10.2118/942107-G

Chang, H., & Zhang, D. (2015). Jointly updating the mean size and spatial distribution of facies in reservoir history matching. *Computational Geosciences*, *19*(4), 727–746. https://doi.org/10.1007/s10596-015-9478-7

Collobert, R., & Weston, J. (2008). A unified architecture for natural language processing: Deep neural networks with multitask learning. *Proceedings of the 25th International Conference on Machine Learning*, 160–167. https://doi.org/10.1145/1390156.1390177

Daubechies, I. (1992). *Ten lectures on wavelets*. SIAM.

E, W., & Yu, B. (2017). The Deep Ritz method: A deep learning-based numerical algorithm for solving variational problems. *ArXiv:1710.00211 [Cs, Stat]*. http://arxiv.org/abs/1710.00211

Fioretto, F., Van Hentenryck, P., Mak, T. W., Tran, C., Baldo, F., & Lombardi, M. (2020). Lagrangian Duality for Constrained Deep Learning. *ArXiv:2001.09394 [Cs, Stat]*. http://arxiv.org/abs/2001.09394

Goodfellow, I., Bengio, Y., & Courville, A. (2016). *Deep learning*. MIT press.

Hangelbroek, T., & Ron, A. (2010). Nonlinear approximation using Gaussian kernels. *Journal of Functional Analysis*, *259*(1), 203–219. https://doi.org/10.1016/j.jfa.2010.02.001

Harbaugh, A. W. (2005). MODFLOW-2005: The U.S. Geological Survey modular ground-water model--the ground-water flow process. In *MODFLOW-2005: The U.S. Geological Survey modular ground-water model—The ground-water flow process* (USGS Numbered Series No. 6-A16; Techniques and Methods, Vols. 6-A16). https://doi.org/10.3133/tm6A16

He, K., Zhang, X., Ren, S., & Sun, J. (2016). *Deep Residual Learning for Image Recognition*. 770–

778.

https://openaccess.thecvf.com/content_cvpr_2016/html/He_Deep_Residual_Learning_CVPR_2016_paper.html

Jagtap, A. D., Kawaguchi, K., & Karniadakis, G. E. (2020). Adaptive activation functions accelerate convergence in deep and physics-informed neural networks. *Journal of Computational Physics*, *404*, 109136. https://doi.org/10.1016/j.jcp.2019.109136

Jo, H., Son, H., Hwang, H. J., & Kim, E. (2019). Deep Neural Network Approach to Forward-Inverse Problems. *ArXiv:1907.12925 [Cs, Math]*. http://arxiv.org/abs/1907.12925

Karpatne, A., Atluri, G., Faghmous, J., Steinbach, M., Banerjee, A., Ganguly, A., Shekhar, S., Samatova, N., & Kumar, V. (2017). Theory-guided Data Science: A New Paradigm for Scientific Discovery from Data. *IEEE Transactions on Knowledge and Data Engineering*, *29*(10), 2318–2331. https://doi.org/10.1109/TKDE.2017.2720168

Karumuri, S., Tripathy, R., Bilionis, I., & Panchal, J. (2020). Simulator-free solution of high-dimensional stochastic elliptic partial differential equations using deep neural networks. *Journal of Computational Physics*, *404*, 109120. https://doi.org/10.1016/j.jcp.2019.109120

Kharazmi, E., Zhang, Z., & Karniadakis, G. E. (2019). Variational Physics-Informed Neural Networks For Solving Partial Differential Equations. *ArXiv:1912.00873 [Physics, Stat]*. http://arxiv.org/abs/1912.00873

Kharazmi, Ehsan, Zhang, Z., & Karniadakis, G. E. (2020). hp-VPINNs: Variational Physics-Informed Neural Networks With Domain Decomposition. *ArXiv:2003.05385 [Cs, Math]*. http://arxiv.org/abs/2003.05385

Langtangen, H. P., Tveito, A., & Winther, R. (1992). Instability of Buckley-Leverett flow in a heterogeneous medium. *Transport in Porous Media*, *9*(3), 165–185. https://doi.org/10.1007/BF00611965

LeCun, Y., Bengio, Y., & Hinton, G. (2015). Deep learning. *Nature*, *521*(7553), 436–444. https://doi.org/10.1038/nature14539

Mo, S., Zhu, Y., Zabaras, N., Shi, X., & Wu, J. (2019). Deep Convolutional Encoder-Decoder Networks for Uncertainty Quantification of Dynamic Multiphase Flow in Heterogeneous Media. *Water Resources Research*, *55*(1), 703–728. https://doi.org/10.1029/2018WR023528

Pang, G., Lu, L., & Karniadakis, G. E. (2018). fPINNs: Fractional Physics-Informed Neural

Networks. *ArXiv:1811.08967 [Physics]*. http://arxiv.org/abs/1811.08967

Paszke, A., Gross, S., Chintala, S., Chanan, G., Yang, E., DeVito, Z., Lin, Z., Desmaison, A., Antiga, L., & Lerer, A. (2017). *Automatic differentiation in pytorch*.

Raissi, M., Perdikaris, P., & Karniadakis, G. E. (2019). Physics-informed neural networks: A deep learning framework for solving forward and inverse problems involving nonlinear partial differential equations. *Journal of Computational Physics*, *378*, 686–707. https://doi.org/10.1016/j.jcp.2018.10.045

Raissi, Maziar, & Karniadakis, G. E. (2018). Hidden physics models: Machine learning of nonlinear partial differential equations. *Journal of Computational Physics*, *357*, 125–141. https://doi.org/10.1016/j.jcp.2017.11.039

Reinbold, P. A. K., Gurevich, D. R., & Grigoriev, R. O. (2020). Using noisy or incomplete data to discover models of spatiotemporal dynamics. *Physical Review E*, *101*(1), 010203. https://doi.org/10.1103/PhysRevE.101.010203

Rong, M., Zhang, D., & Wang, N. (2020). A Lagrangian Dual-based Theory-guided Deep Neural Network. *ArXiv:2008.10159 [Cs, Math, Stat]*. http://arxiv.org/abs/2008.10159

Sallab, A. E., Abdou, M., Perot, E., & Yogamani, S. (2017). Deep Reinforcement Learning framework for Autonomous Driving. *Electronic Imaging*, *2017*(19), 70–76. https://doi.org/10.2352/ISSN.2470-1173.2017.19.AVM-023

Scarpiniti, M., Comminiello, D., Parisi, R., & Uncini, A. (2013). Nonlinear spline adaptive filtering. *Signal Processing*, *93*(4), 772–783. https://doi.org/10.1016/j.sigpro.2012.09.021

Sirignano, J., & Spiliopoulos, K. (2018). DGM: A deep learning algorithm for solving partial differential equations. *Journal of Computational Physics*, *375*, 1339–1364. https://doi.org/10.1016/j.jcp.2018.08.029

Wang, N., Zhang, D., Chang, H., & Li, H. (2020). Deep learning of subsurface flow via theory-guided neural network. *Journal of Hydrology*, *584*, 124700. https://doi.org/10.1016/j.jhydrol.2020.124700

Xu, R. (2020). *Data for Paper Entitled "Weak Form Theory-guided Neural Network (TgNN-wf) for Deep Learning of Subsurface Single and Two-phase Flow."* https://doi.org/10.6084/m9.figshare.12923627.v1

Zhang, D., Li, L., & Tchelepi, H. A. (2000). Stochastic Formulation for Uncertainty Analysis of Two-Phase Flow in Heterogeneous Reservoirs. *SPE Journal*, *5*(01), 60–70.

https://doi.org/10.2118/59802-PA

Zhang, D., & Lu, Z. (2004). An efficient, high-order perturbation approach for flow in random porous media via Karhunen–Loève and polynomial expansions. *Journal of Computational Physics*, *194*(2), 773–794. https://doi.org/10.1016/j.jcp.2003.09.015

Zhang, D., & Tchelepi, H. (1999). Stochastic Analysis of Immiscible Two-Phase Flow in Heterogeneous Media. *SPE Journal*, *4*(04), 380–388. https://doi.org/10.2118/59250-PA

Zhao, H., Liu, F., Li, L., & Luo, C. (2018). A novel softplus linear unit for deep convolutional neural networks. *Applied Intelligence*, *48*(7), 1707–1720. https://doi.org/10.1007/s10489-017-1028-7

Zhu, Y., Zabaras, N., Koutsourelakis, P.-S., & Perdikaris, P. (2019). Physics-constrained deep learning for high-dimensional surrogate modeling and uncertainty quantification without labeled data. *Journal of Computational Physics*, *394*, 56–81. https://doi.org/10.1016/j.jcp.2019.05.024